\pdfoutput=1
\documentclass[11pt]{article}
\usepackage[]{acl}
\usepackage{times}
\usepackage{latexsym}
\usepackage[T5]{fontenc}
\usepackage[utf8]{inputenc}

\usepackage{microtype}
\usepackage{inconsolata}
\usepackage[longtable]{multirow}
\usepackage{multirow}
\usepackage{graphicx}
\usepackage{array}
\usepackage{pifont}
\usepackage{xcolor}
\usepackage{longtable}
\usepackage{amsmath}
\usepackage{rotating}
\usepackage{arydshln}

\title{ViLexNorm: A Lexical Normalization Corpus \\for Vietnamese Social Media Text}

\author{Thanh-Nhi Nguyen\thanks{*Equal contribution.} 
        \and Thanh-Phong Le\footnotemark[1]
        \and Kiet Van Nguyen \\
  Faculty of Information Science and Engineering, \\University of Information Technology, Ho Chi Minh City, Vietnam \\
  Vietnam National University, Ho Chi Minh City, Vietnam \\
  \texttt{\{21521232, 21520395\}@gm.uit.edu.vn},
  \texttt{kietnv@uit.edu.vn} \\}

\begin{document}
\maketitle
\begin{abstract}

Lexical normalization, a fundamental task in Natural Language Processing (NLP), involves the transformation of words into their canonical forms. This process has been proven to benefit various downstream NLP tasks greatly. In this work, we introduce \textbf{Vi}etnamese \textbf{Lex}ical \textbf{Norm}alization (\textsc{ViLexNorm}), the first-ever corpus developed for the Vietnamese lexical normalization task. The corpus comprises over 10,000 pairs of sentences meticulously annotated by human annotators, sourced from public comments on Vietnam's most popular social media platforms. Various methods were used to evaluate our corpus, and the best-performing system achieved a result of 57.74\% using the Error Reduction Rate (ERR) metric \citep{van-der-goot-2019-monoise} with the Leave-As-Is (LAI) baseline. For extrinsic evaluation, employing the model trained on \textsc{ViLexNorm} demonstrates the positive impact of the Vietnamese lexical normalization task on other NLP tasks. Our corpus is publicly available exclusively for research purposes\footnote{\url{https://github.com/ngxtnhi/ViLexNorm}}.

\textbf{Disclaimer:} This paper contains real comments with explicit or potentially sensitive content.

\end{abstract}

\section{Introduction}
In 2022, there were more than 72 million users of social networks in Vietnam, accounting for approximately 73.7\% of the total population\footnote{\url{https://www.statista.com/statistics/278341/number-of-social-network-users-in-selected-countries/}}. The rapid growth of social media has resulted in a significant increase in the volume of data exchanged over the Internet. However, because the data is spontaneous, it naturally contains a wide range of linguistic variances, both intended (e.g., slang, leet speak, puns) and unintended (e.g., mistakes).

\begin{figure}[]
\centering
\includegraphics[width=0.48\textwidth]{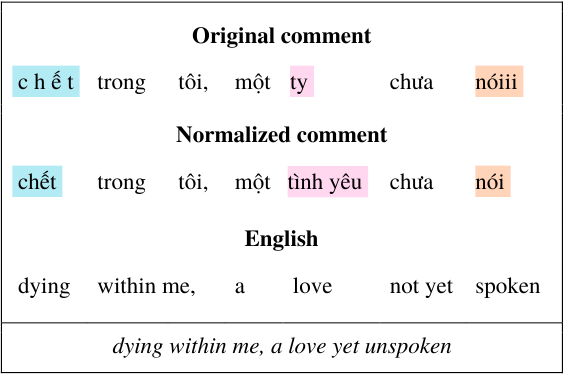}
\caption{Example normalization of “c h ế t trong tôi, một ty chưa nóiii”.}
\label{fig:ex}
\end{figure}

This presents significant challenges for natural language processing software (e.g., \citealp{baldwin-etal-2013-noisy}; \citealp{eisenstein-2013-bad}), which is primarily aimed at analyzing canonical text. One possible approach to enhance the performance of these systems is to normalize the text, thereby increasing its resemblance to the data that NLP systems were originally developed and trained. This task is also called lexical normalization; see Figure~\ref{fig:ex} for the normalization of “c h ế t trong tôi, một ty chưa nóiii” (\textbf{English}: \textit{dying within me, a love yet unspoken}).

In this paper, we define our task of lexical normalization by \citet{van-der-goot-etal-2021-multilexnorm}, expressed by the following formulation:
\renewcommand{\arraystretch}{1.3}
\begin{table}[h]
\begin{tabular}{p{0.95\linewidth}}
\hline
\textbf{Definition - Lexical Normalization}\\
\hline
\textit{Lexical normalization is the task of transforming an utterance into its standard form, word by word, including both one-to-many (1-n) and many-to-one (n-1) replacements.} \\
\hline
\end{tabular}
\label{definition}
\end{table}

In other words, throughout this paper, out-of-vocabulary (OOV) and wrong in-vocabulary (IV) tokens can be normalized to their standard lexical forms and their in-vocabulary counterpart's lexical items, respectively.

The lexical normalization task has been extensively studied in various languages; however, research specific to Vietnamese, a low-resource language, is notably lacking. Recognizing the urgent need for the early-stage exploration of lexical normalization for Vietnamese, we have painstakingly created a corpus named \textsc{ViLexNorm}, encompassing both OOV and IV replacements. We hope this work will serve as a catalyst, encouraging further initiatives to tackle this crucial task for the Vietnamese language.

Our principal contributions in this study consist of the following:
\begin{enumerate}
  \item The establishment of \textsc{ViLexNorm}, the initial corpus for Vietnamese social media data normalization, which encompasses 10,467 sentence pairs. Additionally, we provide a detailed description of our rigorous annotation process. Corpus analysis was thoroughly conducted to grasp the noteworthy phenomena of Vietnamese observed in the domain of social media.
  \item The implementation of two approaches to evaluate the efficacy of our corpus, including \textit{Pre-transformer Models} and \textit{Transformer-based Models}. Interestingly, the pre-trained model for Vietnamese achieved the highest performance along with the relatively competitive performance of the vanilla Transformer, especially considering that it was trained from scratch.
  \item The extrinsic evaluation conducted on various downstream NLP tasks highlights how efficient the Vietnamese lexical normalization task is in improving these tasks' performance.
\end{enumerate}

\section{Related Work} \label{Related Work}

The landscape of lexical normalization research has witnessed significant growth and diversification across various languages over the past decade. This section provides an overview of the foundational work in English and extends to include developments in languages other than English, highlighting the emergence of corpora, advancements in normalization systems, and the downstream impact of lexical normalization on diverse NLP tasks.

Since the foundational work of \citet{han-baldwin-2011-lexical} with LexNorm1.1 a decade ago, lexical normalization has sparked interest in English and several other languages. In the realm of English, the task was followed by subsequent corpora such as LexNorm1.2 \citep{yang-eisenstein-2013-log} and LexNorm2015 \citep{baldwin-etal-2015-shared}. Moving to languages other than English, several corpora were established. Croatian saw the creation of ReLDI-NormTagNER-hr 2.0 \citep{Croatian-2017}, while Serbian had ReLDI-NormTagNER-sr 2.0 \citep{Serbian-2017}. Slovenian, too, had its representation with Janes-Tag 2.0 \citep{Slovenian-2017}. Danish was addressed by the development of DaN+ \citep{plank-etal-2020-dan}. Italian also had a dataset introduced by \citet{van-der-goot-etal-2020-norm}. Shifting the focus to Asian languages, \citet{higashiyama-etal-2021-user} introduced a notable corpus for Japanese. Additionally, \citet{barik-etal-2019-normalization} presented a corpus for code-mixed Indonesian-English, and \citet{makhija-etal-2020-hinglishnorm} developed HinglishNorm for code-mixed Hindi-English. Remarkably, a shared task on multilingual lexical normalization (\textsc{MultiLexNorm} by \citealp{van-der-goot-etal-2021-multilexnorm}) has provided a benchmark including 12 language variants.

Alongside the establishment of corpora, advancements in normalization systems, as exemplified by MoNoise by \citealp{van-der-goot-2019-monoise} and \citealp{muller-etal-2019-enhancing}, have showcased promising outcomes. Furthermore, lexical normalization has been demonstrated to boost various downstream NLP tasks, such as named entity recognition \citep{plank-etal-2020-dan}, POS tagging \citep{pos}, dependency and constituency parsing \citep{van-der-goot-etal-2020-norm}, sentiment analysis \citep{sidorenko2019sentiment}, and machine translation \citep{bhat-etal-2018-universal}.

However, the studies have yet to be applied to Vietnamese. Research efforts have primarily focused on the detection and correction of Vietnamese spelling errors (e.g., \citealp{nguyen2015normalization}; \citealp{nguyen2016text}; \citealp{do2021vsec}; \citealp{vinai}), which are mostly unintended. To the best of our knowledge, \textsc{ViLexNorm} stands as the first work to examine both advertent and inadvertent variations in spelling, encompassing all classifications outlined by \citet{van-der-goot-etal-2018-taxonomy} except phrasal abbreviations.

\section{Corpus Creation} \label{corpus Creation}

\begin{figure*}[]
    \centering
    \includegraphics[width=\textwidth]{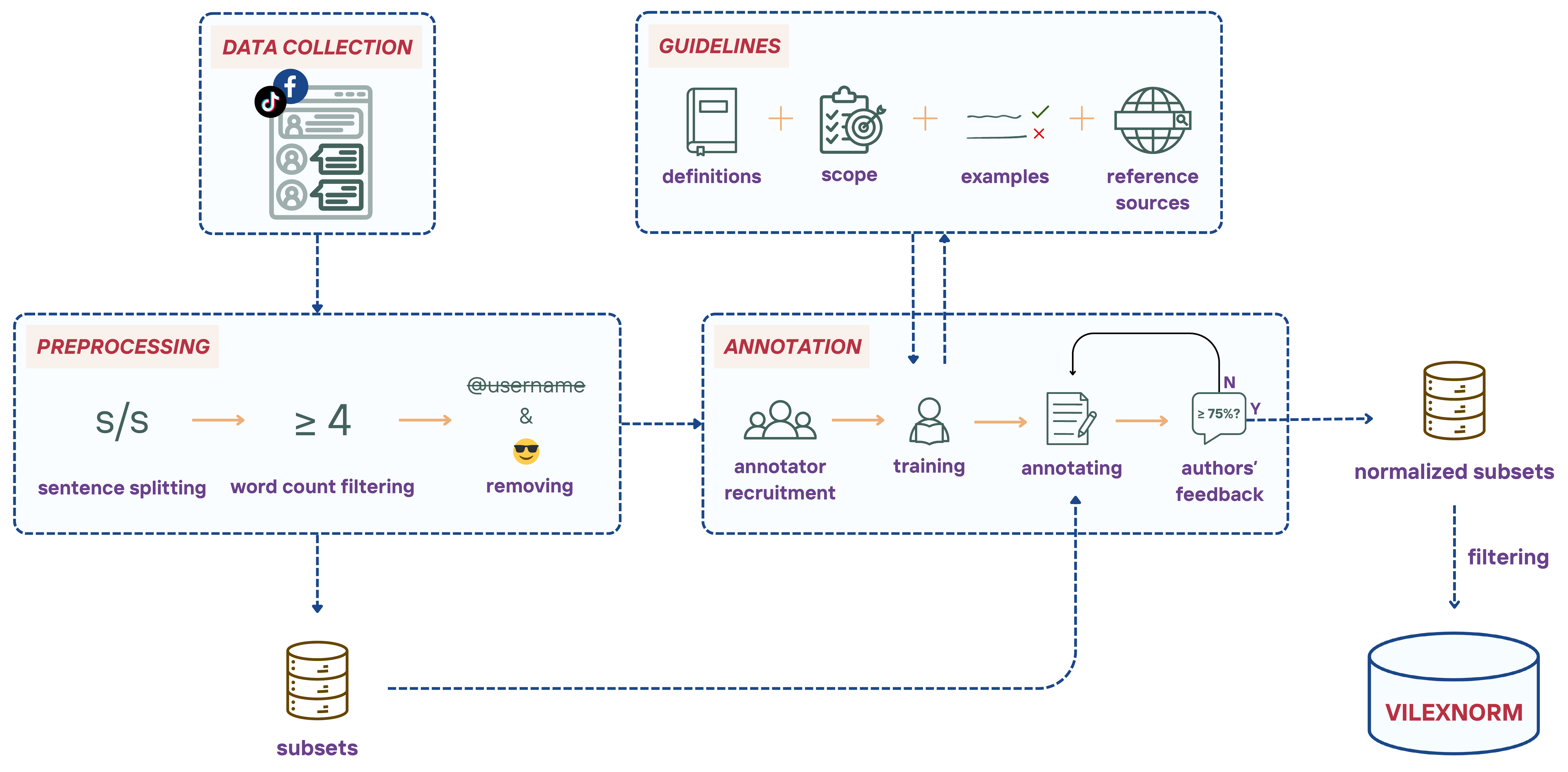}
    \caption{The overview process of creating \textsc{ViLexNorm}.}
    \label{fig:process}
\end{figure*}

In this section, we illustrate our corpus creation. The overview process is depicted in Figure~\ref{fig:process}. 

\subsection{Data Collection and Pre-processing}

Data collection was conducted on two well-known social media platforms including Facebook and TikTok, due to their wide usage and popularity among Vietnamese users\footnote{Statistics sourced from\\ \url{https://www.similarweb.com/top-websites/vietnam/computers-electronics-and-technology/social-networks-and-online-communities/} in May 2023}.

We deliberately picked a wide range of content categories and exclusively included comments from highly engaging public posts. This strategy aimed to amplify the richness and variety of the Vietnamese language expressed across social media. During the pre-processing stage, we divided the comments in paragraph form into individual sentences. Subsequently, we filtered out sentences with fewer than four words to maintain a reasonable annotation density and optimize the annotation process. Furthermore, all usernames in the comments were removed to ensure anonymity. Any emoji characters present in the sentences were also eliminated. In order to avoid overlooking social meaning, as pointed out by \citet{nguyen-etal-2021-learning} and capture social phenomena, we retained all comments that might include toxic or offensive content, and all annotators were fully aware of that.

\subsection{Annotation Process} \label{annotation process}

\textbf{Annotator Recruitment}\ \ \ \ The annotation process involved six native Vietnamese speakers, including two of the authors, encompassing both male and female individuals aged between 20 and 22. The annotators possess extensive familiarity with a wide range of diverse social media platforms and exhibit university entrance examination results in literature surpassing 8.0 on a scale of 10. Furthermore, their academic backgrounds are diverse, spanning fields such as computer science, Vietnamese studies, economics, and construction, contributing to a broad spectrum of perspectives during the annotation process. \\
\\
\textbf{Annotation Guidelines}\ \ As already stated, our objective was to engage annotators from various backgrounds to ensure diverse language perspectives. Consequently, we constructed guidelines encompassing comprehensive definitions of related terms in the task, such as \textit{Vietnamese word}, \textit{non-canonical sentence}, and the annotator's role. This strategy aimed to facilitate a clear understanding of the annotating task. We explicitly outlined the scope of lexical normalization and presented illustrative examples that demonstrated how to normalize each case and common mistakes correctly. In cases where difficulties arose, annotators were recommended to consult reputable resources\footnote{We utilized Tra Tu (a free, open online professional Vietnamese dictionary) and Google for this purpose.}. Furthermore, annotators were encouraged to provide suggestions to enhance the feasibility of the guidelines. \\
\\
\textbf{Training Phase}\ \ In the initial phase of the annotation stage, the annotators were provided with guidelines and underwent a training session. They were assigned to a subset of 100 sentences and asked to estimate the number of subsets they could annotate in a day. We allowed the annotators to freely determine their workload in order to ensure the annotations' quality. \\
\\
\textbf{Main Annotation}\ \ For each annotated subset of 100 sentences, the annotators received feedback on a sample of 20 random sentences from the authors. We calculated the percentage of sentences for which the authors agreed on the normalization, specifically when they mutually agreed that the sentence was completely normalized. If the agreement score between the annotator's annotations and the authors' annotations was below 75\%, the annotator was requested to re-annotate the entire subset. Notably, agreement between subsets annotated by either of the two authors was evaluated by the other author.

Throughout both the Training and Main Annotation phases, no subset required re-annotation more than once; thus, no annotators were eliminated. \\
\\
\textbf{Inter-annotator Agreements} The agreement between annotators was averaged across all subsets during the main annotation phase. Additionally, as the authors were involved in the annotation task, the agreement between them was computed separately. The averaged inter-annotator agreement for all subsets during the main annotation phase was 88.46\% between the authors and other annotators and 93.54\% between the two authors. The observed scores reflect a high level of concordance between annotations, demonstrating strong agreement between the annotators and the authors in our task. \\
\\
\textbf{Filtering}\ \ Following the main annotation phase, we excluded sentences that did not contain any words requiring normalization in our defined scope. Afterward, the \textsc{ViLexNorm} corpus comprises a total of 10,467 pairs of sentences.

\subsection{Corpus Statistics}

The corpus \textsc{ViLexNorm} consists of 10,467 comment pairs following the annotation process. These are further partitioned into three subsets: training, development, and test, distributed in an 8:1:1 ratio. The corpus encompasses a total of 20,061 word pairs, comprising a total of 3,489 distinct pairs derived from the comments.

Vietnamese is a monosyllabic language wherein every syllable is distinctly separated by a space in its written form. Alternatively, a word in Vietnamese can consist of multiple syllables separated by spaces. This separation aids in proper pronunciation and comprehension of words, reflecting the Vietnamese language's unique phonological and orthographic features. In light of this, we undertook an analysis of \textsc{ViLexNorm} by considering the element of syllable counts, aiming to delve into the distinctive characteristics of Vietnamese.

A thorough distribution analysis of the words is provided in Table~\ref{tab:stas}. It is indisputable that the majority of Vietnamese individuals utilize 1-syllable canonical words with the utmost frequency when engaging on various social media platforms. Moreover, we observe a noteworthy pattern in the 2-syllable and 3-syllable categories. The count of normalized words (2,741 for 2-syllable and 104 for 3-syllable) surpasses the count of non-canonical words (396 for 2-syllable and 7 for 3-syllable), suggesting that individuals deliberately opt for shorter variations of words when communicating through online channels. This inclination towards brevity and efficiency in conveying messages aligns with the typical characteristics of online discourse.

\renewcommand{\arraystretch}{1.2}
\begin{table*}[]
\centering
\resizebox{.7\textwidth}{!}{\begin{tabular}{r>{\centering\arraybackslash}p{2cm}>{\centering\arraybackslash}p{2cm}|>{\centering\arraybackslash}p{2cm}>{\centering\arraybackslash}p{2cm}} 
\hline
\multicolumn{1}{l}{\multirow{2}{*}{\textbf{Number of Syllables}}} & \multicolumn{2}{c}{\textbf{Non-canonical Words}} & \multicolumn{2}{c}{\textbf{Normalized Words}} \\ 
\cline{2-5}
\multicolumn{1}{l}{} & \textit{Total} & \multicolumn{1}{c}{\textit{Distinct}} & \textit{Total} & \textit{Distinct} \\ 
\hline
\textit{1} & 19,658 & 3,188 & 17,207 & 2,707 \\
\textit{2} & 396 & 295 & 2,741 & 736 \\
\textit{3} & 7 & 6 & 104 & 41 \\
\textit{4} &-  &-  & 9 & 5 \\
\textit{Total} & 20,061 & 3,489 & 20,061 & 3,489 \\
\hline
\end{tabular}}
\caption{\textsc{ViLexNorm} statistics showing the number of words categorized by syllable count. \textbf{Non-canonical Words} refers to words found in the original sentences that needed normalization. \textbf{Normalized Words} represents the count of words normalized from their non-canonical forms. \textit{Total} denotes the total count of words, and \textit{Distinct} signifies the count of distinct words.}
\label{tab:stas}
\end{table*}

To assess the extent of linguistic diversity observed on social networks, we conducted an analysis of the standard words that displayed the highest number of variations, as depicted in Table~\ref{tab:var}. The results yielded fascinating statistics. For example, the word "không" (\textit{no}) demonstrated an impressive total of 53 variations, which underscores the creative language used by Vietnamese individuals in the online sphere. Additionally, we explored the top ten most frequently normalized terms, detailed in Appendix \ref{sec:appendixa}.

\begin{table}
\small
\centering
\begin{tabular}{lr} 
\hline
\textbf{Standard word} & \multicolumn{1}{l}{\textbf{Number of variants}} \\ 
\hline
không (no) & 53 \\
rồi (already) & 50 \\
vậy (so) & 34 \\
quá (very) & 34 \\
thôi (stop) & 33 \\
ơi (hey) & 31 \\
biết (know) & 24 \\
trời (god) & 23 \\
được (okay) & 22 \\
đi (go) & 21 \\
\hline
\end{tabular}
\caption{Standard words with the most variations in \textsc{ViLexNorm}.}
\label{tab:var}
\end{table}

\section{Intrinsic Evaluation} \label{Intrinsic Evaluation}

This section focuses on the intrinsic evaluation of \textsc{ViLexNorm}, examining its empirical performance through diverse experiments and methodologies. We explore methods ranging from pre-transformer structures to transformer-based structures in the lexical normalization task. Subsequently, we outline the experimental setup, including data configurations, training procedures, and metrics. Finally, we present evaluation results, analyzing each method's performance and offering insights into the efficiency and effectiveness of \textsc{ViLexNorm}.

\subsection{Methods}

To establish empirical performances on \textsc{ViLexNorm}, we conducted various experiments using different methods:

\begin{itemize}
    \item \textbf{Pre-transformer Structures:}
    We initiated by employing well-established architectures predating the widespread adoption of transformer-based models in NLP tasks. This category includes Long Short-Term Memory (LSTM; \citealp{lstm}) and Bidirectional Gated Recurrent Units (BiGRU; \citealp{cho-etal-2014-properties}) with Attention mechanism \citep{bahdanau2014neural}. We chose these architectures due to their proven effectiveness in sequence modeling and their historical prominence in NLP tasks.

    \item \textbf{Transformer-based Structures:}
    We further delved into transformer-based structures, including training of vanilla Transformer from scratch \citep{vaswani2017attention} and fine-tuning BARTpho \citep{bartpho}, a pre-trained Sequence-to-Sequence model for Vietnamese. These selections were motivated by the rapid advancements in deep learning, with the anticipation that they would optimize task performance.
\end{itemize}

\subsection{Experimental Setup}

In this setup, we approached the lexical normalization task as a sequence-to-sequence problem, where the input comprised a sentence containing at least one word in its unnormalized form, and the objective was to generate the corresponding normalized sentence. Except for BARTpho, which inherently provides options for syllable-level and word-level input, we assessed the models on both segmented and unsegmented versions of the corpus using VnCoreNLP \citep{vu-etal-2018-vncorenlp} to understand the influence of word segmentation on their performance. Additionally, we applied Byte-Pair encoder \citep{sennrich-etal-2016-neural} with a vocabulary size of 7000.

For the BiGRU and LSTM models, the model training commenced with a batch size of 32, employing the Adam optimizer along with cross-entropy loss. The training spanned 40 epochs, utilizing a learning rate of 0.01. The same experimental setup was applied to the vanilla Transformer, albeit with a learning rate of 0.0001. We explored both versions of BARTpho, namely BARTpho\textsubscript{syllable} and BARTpho\textsubscript{word}, publicly available on Hugging Face\footnote{\url{https://huggingface.co/vinai}}. Within this method, we designated the epoch count as 10, utilizing a learning rate of 5e-5.

We utilized a system with 13GB RAM and an NVIDIA Tesla T4 GPU to train all initial models. The manual seed for BARTpho was set to 42, whereas for the remaining models, it was established as 0. This was done to ensure reproducibility and consistency in the results.

\subsection{Metrics}

This paper employed the Error Reduction Rate (ERR) proposed by \citet{van-der-goot-2019-monoise} as the primary metric. ERR assesses the reduction in errors compared to a previous model and serves as a normalized measure of token-level accuracy, considering the percentage of tokens requiring normalization. Since there is currently no standard normalization model for Vietnamese, the Leave-As-Is (LAI) baseline, which retains the input word, was utilized.

The ERR formula is as follows: 
\begin{align}
\text{{ERR}} = \frac{{\text{{Accuracy}$_{system}$} - \text{{Accuracy}$_{baseline}$}}}{{1.0 - \text{{Accuracy}$_{baseline}$}}}
\end{align}

The ERR typically falls within the range of 0.0 to 1.0, whereby a negative ERR suggests more incorrect token normalizations than correct ones. It is worth noting that the Leave-As-Is baseline, which returns the input words without any alterations, will inevitably produce an ERR value of 0.0.

In the context of 1-n and n-1 transformations, we utilize the Levenshtein distance metric \cite{levenshtein1966binary} to calculate accuracy at the token level.

As stated by \citet{van2019normalization}, ERR has the limitation of not providing insight into the distinction between false positives (FP) and false negatives (FN). This metric does not inform us whether the system normalizes excessively or cautiously. Therefore, we also incorporated two additional metrics: Precision and Recall.

\subsection{Evaluation Results}

Table~\ref{tab:res} displays the intrinsic evaluation results for various methods regarding Error Reduction Rate (ERR), Precision, and Recall.

\renewcommand{\arraystretch}{1.5}
\begin{table*}[]
\small
\centering
\resizebox{.7\textwidth}{!}{\begin{tabular}{lllccc} 
\hline
 & \textbf{Method} & \textbf{Level} & \textbf{ERR} & \textbf{Precision} & \textbf{Recall} \\ 
\hline
\multirow{4}{*}{\textbf{Pre-transformer structures}} & \multirow{2}{*}{\textit{LSTM}} & Syllable & -4.3781 & 0.1178 & 0.1187 \\
 &  & Word & -4.1319 & 0.1225 & 0.1222 \\ 
\cline{2-6}
 & \multirow{2}{*}{\textit{BiGRU + Attention}} & Syllable & -0.2483 & 0.8350 & 0.8369 \\
 &  & Word & -0.3025 & 0.8182 & 0.8015 \\ 
\hline
\multirow{4}{*}{\textbf{Transformer-based structures}} & \multirow{2}{*}{\textit{Vanilla Transformer}} & Syllable & \textbf{0.3394} & \textbf{0.9090} & \textbf{0.9104} \\
 &  & Word & 0.2903 & 0.8944 & 0.8950 \\ 
\cline{2-6}
 & \textit{BARTpho\textsubscript{syllable}} & Syllable & \textbf{0.5774} & \textbf{0.9332} & \textbf{0.9193} \\
 & \textit{BARTpho\textsubscript{word}} & Word & 0.2269 & 0.8912 & 0.8735 \\
\hline
\end{tabular}}
\caption{Intrinsic evaluation of models trained on \textsc{ViLexNorm}, showcasing Error Reduction Rate (ERR), Precision, and Recall. Results are presented across pre-transformer and transformer-based architectures, considering both word and syllable-level data configurations.}
\label{tab:res}
\end{table*}

In terms of pre-transformer structures, using LSTM with both data versions resulted in ERR values of -4.3781 and -4.1319, respectively. These negative ERR values indicate that the models had a higher error rate than the baseline LAI approach. However, transitioning to BiGRU with the Attention mechanism showed improvement, bringing ERR closer to zero, with -0.2483 for syllable level and -0.3025 for word level. Notably, BiGRU achieved positive precision and recall of around 0.80 to 0.84.

Moving to transformer-based structures, the vanilla Transformer displayed intriguing results, achieving an ERR of 0.3394, a precision of 0.9090, and a recall of 0.9104 for the syllable version of data. Remarkably, the BARTpho\textsubscript{syllable} model showcased a significant positive ERR of 0.5774, emphasizing its capacity to substantially reduce errors and enhance both precision (0.9332) and recall (0.9193). For the word-level data, the vanilla Transformer and BARTpho\textsubscript{word} also displayed improvement over the LAI baseline, achieving ERRs of 0.2903 and 0.2269, respectively. However, this improvement was less pronounced compared to their syllable-level counterparts. These outcomes underscore that transformer-based structures perform exceptionally well, even without the necessity of word segmentation, reaffirming their alignment with Vietnamese linguistic features and suggesting an enhanced capability to capture and process these linguistic nuances.

Overall, despite encountering challenges with pre-transformer structures resulting in higher error rates than the baseline, the advancements observed with transformer-based architectures, particularly BARTpho\textsubscript{syllable}, demonstrate potential for substantial error reduction, offering an encouraging outlook for further advancements in the lexical normalization task for Vietnamese.

\subsection{Effects of Non-canonical Word Ratio in Sentences on Normalization Efficiency}

In order to gain insights into how the ratio of words necessitating normalization within a sentence affects the efficiency of the normalization process, we conducted a thorough analysis on the development set using the ERR score of BARTpho\textsubscript{syllable} due to its superior performance.

\begin{figure}[h]
\centering
\includegraphics[width=0.5\textwidth]{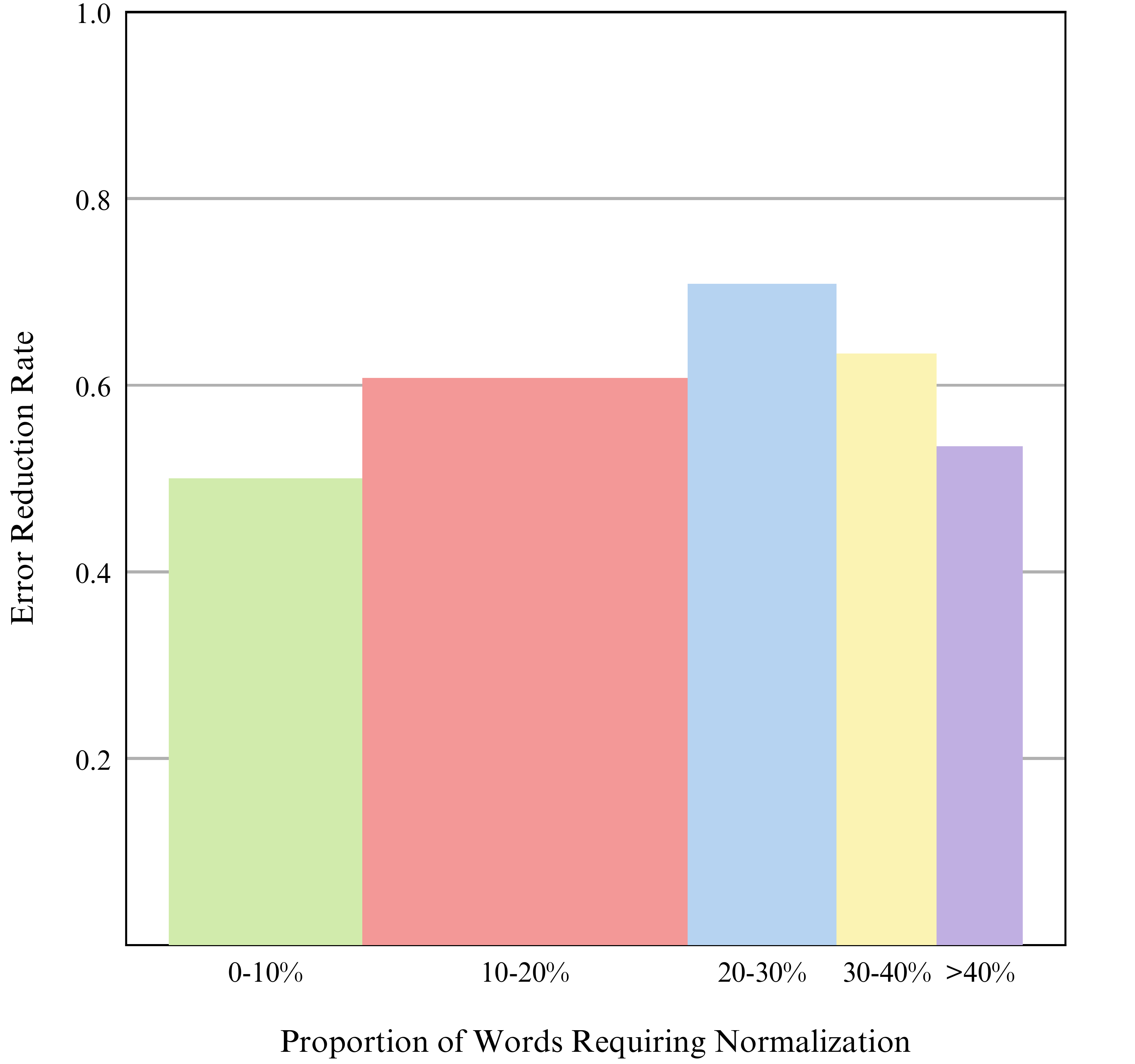}
\caption{Performance analysis of BARTpho\textsubscript{syllable} on the development set of \textsc{ViLexNorm}, demonstrating an association between non-canonical word ratio and normalization efficiency.
}
\label{fig:analysis}
\end{figure}

Figure \ref{fig:analysis} provides a graphical insight into the relationship between non-canonical word ratios and the corresponding ERR performances. The width of the columns is proportional to the number of samples in each category.

The ERR performance followed a distinct pattern with respect to the ratio of words requiring normalization. Specifically, the normalization efficiency appeared to improve as the ratio of words to be normalized increased, peaking in the range of 20-30\%. Beyond this range, the efficiency slightly decreased, though it remained higher than the 0-10\% and 10-20\% categories.

This pattern suggests that sentences with a moderate proportion of words needing normalization (20-30\%) are optimally suited for the normalization process. The normalization system may have been effectively trained and fine-tuned to handle this range, resulting in enhanced efficiency. However, as the ratio of words needing normalization exceeds this range, the system encounters challenges, potentially due to increased linguistic complexity or noise within the sentence.

\section{Extrinsic Evaluation} \label{Extrinsic Evaluation}

This section extends the assessment of \textsc{ViLexNorm} beyond intrinsic measures, exploring its impact on downstream NLP tasks. Through experiments, we investigate how the normalization system enhances performance in emotion recognition, hate speech detection, and spam detection. We also assess its efficacy in scenarios without Vietnamese diacritics, providing insights into its adaptability and real-world effectiveness.

\subsection{The Impact of Lexical Normalization on Downstream NLP Tasks Performance} \label{ex1}

To validate our normalization system's practical applicability and effectiveness, we conducted extrinsic evaluations across three specific tasks. These tasks consisted of emotion recognition using the UIT-VSMEC corpus \citep{vsmec}, hate speech detection using the ViHSD dataset  \cite{vihsd}, and spam detection using the ViSPAM dataset \citep{vispam}. The UIT-VSMEC corpus comprises 6,927 sentences from Facebook, categorized into seven emotion labels through human annotation. Conversely, the ViHSD dataset, consisting of 33,400 comments, was annotated into three labels specifically for hate speech detection on various social networking platforms. Lastly, the ViSPAM dataset, with its 19,868 reviews, was curated to identify spam reviews, particularly opinion-based ones, on Vietnamese E-commerce platforms. In our assessment of ViSPAM, we focused solely on the binary classification task, determining whether a review is spam or not. It is important to note that emoji characters were excluded from all three datasets as our normalization system is incapable of handling emojis.

For the extrinsic evaluation, we leveraged a diverse set of models for all three tasks. TextCNN \citep{kim-2014-convolutional}, recognized for its efficiency in text classification, was one of the key models. We also incorporated Bidirectional LSTM (BiLSTM) and Gated Recurrent Unit (GRU), both renowned for their proficiency in sequence modeling. Furthermore, we utilized PhoBERT \citep{nguyen-tuan-nguyen-2020-phobert}, a state-of-the-art monolingual language model pre-trained specifically for Vietnamese, for this evaluation. See Appendix~\ref{sec:appendixc} for details on hyperparameters and training.

 Our chosen normalization system for this evaluation was the BARTpho\textsubscript{syllable} due to its superior performance observed in the intrinsic evaluation. In this setup, we employed the normalized versions of the input texts generated by our chosen normalization system as the input for the models. Notably, we trained the models three times while keeping the normalization model frozen, underlining the effectiveness of our normalization system in enhancing downstream task performance. The averaged results from these experiments are detailed in Table \ref{tab:extrinsic}, providing a comprehensive view of the performance of these models before and after normalization.

\begin{table*}[]
\centering
\begin{tabular}{lcc|cc|cc} 
\hline
\multicolumn{1}{c}{\multirow{2}{*}{~}} & \multicolumn{2}{c}{\textbf{UIT-VSMEC}} & \multicolumn{2}{c}{\textbf{ViHSD}} & \multicolumn{2}{c}{\textbf{ViSPAM}} \\ 
\cline{2-7}
\multicolumn{1}{c}{} & \textit{Before} & \multicolumn{1}{c}{\textit{After}} & \textit{Before} & \multicolumn{1}{c}{\textit{After}} & \textit{Before} & \textit{After} \\ 
\hline
\textbf{TextCNN} & 29.48 & \textbf{29.85} & 57.38 & \textbf{58.92} & 78.29 & \textbf{78.31} \\
\textbf{BiLSTM} & 23.43 & \textbf{25.23} & 58.10 & \textbf{60.88} & 76.93 & \textbf{77.91~} \\
\textbf{GRU} & 27.85 & \textbf{30.10} & 60.92 & \textbf{61.23} & \textbf{79.35} & 78.93 \\
\textbf{PhoBERT} & 59.15 & \textbf{62.03} & 65.91 & \textbf{66.54} & \textbf{89.28} & 88.21 \\
\hline
\end{tabular}
\caption{F1-macro scores of models before and after lexical normalization on three NLP downstream tasks: emotion recognition (UIT-VSMEC), hate speech detection (ViHSD), and spam detection (ViSPAM).}
\label{tab:extrinsic}
\end{table*}

The results demonstrate that the application of our normalization system exhibited improved F1-macro scores in both UIT-VSMEC and ViHSD cases. These findings indicate the potential affirmative impact of our normalization systems on improving emotion recognition and hate speech detection. However, the outcomes for ViSPAM did not exhibit significant promise, showing a slight decrease in half of the cases. This suggests that the binary classification task of identifying spam messages is relatively uncomplicated, enabling models to comprehend essential characteristics without requiring a normalization stage. Another potential reason for this outcome may be attributed to the loss of important features through the normalization of non-standard input, which is crucial for spam detection.

In summary, the extrinsic evaluation strongly affirms that integrating our normalization system enhances input data quality, resulting in improved performance across diverse NLP tasks, especially in complex tasks requiring sophisticated pre-processing strategies, highlighting the versatile applicability of our normalization approach.

\subsection{Normalization Impact when Lacking Vietnamese Diacritics}

Vietnamese diacritics, commonly known as diacritical marks or accents, play a pivotal role in the orthography and semantic interpretation of the language. These diacritics, encompassing tones and additional markers, are indispensable in differentiating words with similar spellings but distinct meanings. For example, the term "ma" can denote "ghost," "mother," "rice seedling," or "which," contingent upon the employed tone. In this section, our objective was to investigate the efficacy of the normalization system, particularly BARTpho\textsubscript{syllable}, in augmenting downstream task efficiency using PhoBERT in the absence of Vietnamese diacritics. We conducted experiments by removing varying percentages of diacritics from each comment in the UIT-VSMEC and ViHSD datasets. The results depicted in Table~\ref{tab:diacritics} showcase the performance of PhoBERT before and after normalization using BARTpho\textsubscript{syllable} under various diacritic removal percentages: 25\%, 50\%, 75\%, and 100\%.

\begin{table*}[]
\centering
\begin{tabular}{lcc|cc|cc|cc} 
\hline
\multirow{2}{*}{\textbf{~}} & \multicolumn{2}{c}{\textbf{25\%}} & \multicolumn{2}{c}{\textbf{50\%}} & \multicolumn{2}{c}{\textbf{75\%}} & \multicolumn{2}{c}{\textbf{100\%}} \\ 
\cline{2-9}
 & \textit{Before} & \multicolumn{1}{c}{\textit{After}} & \textit{Before}\textbf{} & \multicolumn{1}{c}{\textit{After}\textbf{}} & \textit{Before}\textbf{} & \multicolumn{1}{c}{\textit{After}\textbf{}} & \textit{Before}\textbf{} & \textit{After}\textbf{} \\ 
\cline{2-9}
\textbf{UIT-VSMEC} & 53.63 & \textbf{53.94} & 40.79 & \textbf{43.50} & \textbf{32.62} & 29.59 & \textbf{32.77} & 31.91\textbf{} \\
\textbf{ViHSD} & 61.59\textbf{} & \textbf{62.27} & 53.92\textbf{} & \textbf{58.81} & \textbf{57.08} & 56.78\textbf{} & \textbf{57.18} & 56.85\textbf{} \\
\hline
\end{tabular}
\caption{PhoBERT's F1-macro score comparison before and after lexical normalization on UIT-VSMEC (emotion recognition) and ViHSD (hate speech detection) datasets across varying diacritic removal levels (25\%, 50\%, 75\%, 100\%).}
\label{tab:diacritics}
\end{table*}

PhoBERT exhibited a consistent decline in performance as diacritics were removed, compared to the performance discussed in Section~\ref{ex1} where diacritics were retained. This decrease in performance is an anticipated outcome given that diacritics carry essential linguistic information in Vietnamese, and their removal can impact the models' ability to process and understand the text accurately.

Upon examining specific diacritic removal percentages, an interesting pattern emerged. Both datasets, UIT-VSMEC and ViHSD, exhibited an increase in performance after normalization when 25\% and 50\% of diacritics were removed. However, the increase was notably higher at the 50\% removal mark, indicating a more significant impact of normalization at this level.

On the other hand, as the diacritic removal increased to 75\% and 100\%, both datasets demonstrated a decrease in performance after normalization. Interestingly, the F1-macro score before normalization at 100\% diacritic removal surpasses that at 75\%, a surprising observation. This pattern suggests that the near-complete removal of diacritics could introduce additional noise or modify the linguistic context in a manner detrimental to the model's performance, even after normalization. Another plausible factor could be the limited presence of non-diacritic samples in our corpus. Expanding the corpus to include more non-diacritic samples could potentially enhance model performance across varying diacritic removal levels, a direction worth considering in future research.

\section{Conclusion and Future Work} \label{Conclusion}

Our paper introduced \textsc{ViLexNorm}, a novel corpus expressly designed for the lexical normalization task of Vietnamese social media data. The corpus analysis demonstrated captivating characteristics of the Vietnamese language used on social media. We conducted empirical evaluations employing various methods on this corpus, and the BARTpho\textsubscript{syllable} model emerged as the top performer, achieving an impressive ERR score of 57.74\% and a Precision score of 93.32\%. Additionally, we harnessed the potential of \textsc{ViLexNorm} to assess the impact of lexical normalization on downstream NLP tasks, and the results were encouraging. As the pioneering effort in the lexical normalization task for Vietnamese, we hope that our corpus contributes to the diversity of the multilingual lexical normalization task. Furthermore, we expect this work to motivate and inspire further exploration and research in handling noisy data on the Internet, advancing the field of lexical normalization in Vietnamese NLP research.

Promising avenues for advancement in this task are considered for our future research. Our roadmap includes not only expanding the corpus in both scale and diversity but also incorporating a variety of Vietnamese language variants found across the Internet (e.g., text lacking diacritic marks). Additionally, we intend to conduct a thorough analysis of agreement, exploring metrics like Cohen’s kappa score \cite{cohen1960coefficient}, to gain deeper insights into the quality and consistency of the corpus. Moreover, we are inclined towards a comprehensive exploration and adaptation of state-of-the-art models and methods, including MoNoise (\citealp{van-der-goot-2019-monoise}), with the goal of identifying optimal solutions for the lexical normalization task and advancing the development of highly effective multilingual lexical normalization models that can effectively bridge language-specific gaps. Another important aspect of our future work involves expanding the scope of extrinsic evaluations to encompass a broader range of NLP tasks, including dependency parsing and POS tagging (\citealp{van-der-goot-etal-2021-multilexnorm}; \citealp{van2019normalization}). These tasks require label adjustments during normalization due to the monosyllabic nature of Vietnamese, necessitating the investigation of adaptive methods for monosyllabic languages and contributing to a more diverse language landscape in practical language processing scenarios.

\clearpage

\section*{Limitations and Ethical Considerations}

\textbf{Limitations} 

In addition to the mentioned contributions, it is important to acknowledge the presence of several limitations in our work. The \textsc{ViLexNorm} corpus was formed within six months during the research, potentially failing to represent the broader linguistic developments throughout time accurately. Additionally, the presence of incomprehensible comments in our corpus due to the lack of context, showcasing the diverse language used on the Internet, could potentially influence the overall performance in real-world applications. The inter-annotator agreement, while analyzed to some extent, remains relatively shallow, and further exploration is needed to gain a more in-depth understanding of the quality and consistency of our corpus.\\
\\
\textbf{Ethical Considerations}

During the recruitment stage, we clearly informed the annotators that the tasks would involve sensitive and potentially harmful content. The purpose of granting annotators the ability to manage their workload, as mentioned in Section \ref{annotation process}, was to prioritize their mental well-being. If, at any point, the annotators found the annotation tasks to be overwhelming, they were strongly encouraged to notify the authors.
Annotators received compensation of \$0.02 for each comment normalized, which typically required an average duration of 10 seconds to finish.

\section*{Acknowledgements}

We would like to express our gratitude to the annotators for their diligent efforts. Additionally, we extend our sincere appreciation to the reviewers for their valuable feedback and insights, which significantly contributed to the improvement of this paper.

This research was supported by The VNUHCM-University of Information Technology’s Scientific Research Support Fund.

\bibliography{anthology,custom}

\begin{thebibliography}{41}
\expandafter\ifx\csname natexlab\endcsname\relax\def\natexlab#1{#1}\fi

\bibitem[{Bahdanau et~al.(2014)Bahdanau, Cho, and Bengio}]{bahdanau2014neural}
Dzmitry Bahdanau, Kyunghyun Cho, and Yoshua Bengio. 2014.
\newblock Neural machine translation by jointly learning to align and translate.
\newblock \emph{arXiv preprint arXiv:1409.0473}.

\bibitem[{Baldwin et~al.(2013)Baldwin, Cook, Lui, MacKinlay, and Wang}]{baldwin-etal-2013-noisy}
Timothy Baldwin, Paul Cook, Marco Lui, Andrew MacKinlay, and Li~Wang. 2013.
\newblock \href {https://aclanthology.org/I13-1041} {How noisy social media text, how diffrnt social media sources?}
\newblock In \emph{Proceedings of the Sixth International Joint Conference on Natural Language Processing}, pages 356--364, Nagoya, Japan. Asian Federation of Natural Language Processing.

\bibitem[{Baldwin et~al.(2015)Baldwin, de~Marneffe, Han, Kim, Ritter, and Xu}]{baldwin-etal-2015-shared}
Timothy Baldwin, Marie~Catherine de~Marneffe, Bo~Han, Young-Bum Kim, Alan Ritter, and Wei Xu. 2015.
\newblock \href {https://doi.org/10.18653/v1/W15-4319} {Shared tasks of the 2015 workshop on noisy user-generated text: {T}witter lexical normalization and named entity recognition}.
\newblock In \emph{Proceedings of the Workshop on Noisy User-generated Text}, pages 126--135, Beijing, China. Association for Computational Linguistics.

\bibitem[{Barik et~al.(2019)Barik, Mahendra, and Adriani}]{barik-etal-2019-normalization}
Anab~Maulana Barik, Rahmad Mahendra, and Mirna Adriani. 2019.
\newblock \href {https://doi.org/10.18653/v1/D19-5554} {Normalization of {I}ndonesian-{E}nglish code-mixed {T}witter data}.
\newblock In \emph{Proceedings of the 5th Workshop on Noisy User-generated Text (W-NUT 2019)}, pages 417--424, Hong Kong, China. Association for Computational Linguistics.

\bibitem[{Bhat et~al.(2018)Bhat, Bhat, Shrivastava, and Sharma}]{bhat-etal-2018-universal}
Irshad Bhat, Riyaz~A. Bhat, Manish Shrivastava, and Dipti Sharma. 2018.
\newblock \href {https://doi.org/10.18653/v1/N18-1090} {{U}niversal {D}ependency parsing for {H}indi-{E}nglish code-switching}.
\newblock In \emph{Proceedings of the 2018 Conference of the North {A}merican Chapter of the Association for Computational Linguistics: Human Language Technologies, Volume 1 (Long Papers)}, pages 987--998, New Orleans, Louisiana. Association for Computational Linguistics.

\bibitem[{Cho et~al.(2014)Cho, van Merri{\"e}nboer, Bahdanau, and Bengio}]{cho-etal-2014-properties}
Kyunghyun Cho, Bart van Merri{\"e}nboer, Dzmitry Bahdanau, and Yoshua Bengio. 2014.
\newblock \href {https://doi.org/10.3115/v1/W14-4012} {On the properties of neural machine translation: Encoder{--}decoder approaches}.
\newblock In \emph{Proceedings of {SSST}-8, Eighth Workshop on Syntax, Semantics and Structure in Statistical Translation}, pages 103--111, Doha, Qatar. Association for Computational Linguistics.

\bibitem[{Cohen(1960)}]{cohen1960coefficient}
Jacob Cohen. 1960.
\newblock A coefficient of agreement for nominal scales.
\newblock \emph{Educational and psychological measurement}, 20(1):37--46.

\bibitem[{Do et~al.(2021)Do, Nguyen, Bui, and Vo}]{do2021vsec}
Dinh-Truong Do, Ha~Thanh Nguyen, Thang~Ngoc Bui, and Hieu~Dinh Vo. 2021.
\newblock Vsec: Transformer-based model for vietnamese spelling correction.
\newblock In \emph{PRICAI 2021: Trends in Artificial Intelligence: 18th Pacific Rim International Conference on Artificial Intelligence, PRICAI 2021, Hanoi, Vietnam, November 8--12, 2021, Proceedings, Part II 18}, pages 259--272. Springer.

\bibitem[{Eisenstein(2013)}]{eisenstein-2013-bad}
Jacob Eisenstein. 2013.
\newblock \href {https://aclanthology.org/N13-1037} {What to do about bad language on the internet}.
\newblock In \emph{Proceedings of the 2013 Conference of the North {A}merican Chapter of the Association for Computational Linguistics: Human Language Technologies}, pages 359--369, Atlanta, Georgia. Association for Computational Linguistics.

\bibitem[{Erjavec et~al.(2017)Erjavec, Fi{\v{s}}er, {\v{C}}ibej, Arhar~Holdt, Ljube{\v{s}}i{\'c}, and Zupan}]{Slovenian-2017}
Toma{\v{z}} Erjavec, Darja Fi{\v{s}}er, Jaka {\v{C}}ibej, {\v{S}}pela Arhar~Holdt, Nikola Ljube{\v{s}}i{\'c}, and Katja Zupan. 2017.
\newblock Cmc training corpus janes-tag 2.0.

\bibitem[{Han and Baldwin(2011)}]{han-baldwin-2011-lexical}
Bo~Han and Timothy Baldwin. 2011.
\newblock \href {https://aclanthology.org/P11-1038} {Lexical normalisation of short text messages: Makn sens a {\#}twitter}.
\newblock In \emph{Proceedings of the 49th Annual Meeting of the Association for Computational Linguistics: Human Language Technologies}, pages 368--378, Portland, Oregon, USA. Association for Computational Linguistics.

\bibitem[{Higashiyama et~al.(2021)Higashiyama, Utiyama, Watanabe, and Sumita}]{higashiyama-etal-2021-user}
Shohei Higashiyama, Masao Utiyama, Taro Watanabe, and Eiichiro Sumita. 2021.
\newblock \href {https://doi.org/10.18653/v1/2021.naacl-main.438} {User-generated text corpus for evaluating {J}apanese morphological analysis and lexical normalization}.
\newblock In \emph{Proceedings of the 2021 Conference of the North American Chapter of the Association for Computational Linguistics: Human Language Technologies}, pages 5532--5541, Online. Association for Computational Linguistics.

\bibitem[{Ho et~al.(2019)Ho, Nguyen, Nguyen, Pham, Nguyen, Nguyen, and Nguyen}]{vsmec}
Vong~Anh Ho, Duong~Huynh{-}Cong Nguyen, Danh~Hoang Nguyen, Linh~Thi{-}Van Pham, Duc{-}Vu Nguyen, Kiet~Van Nguyen, and Ngan~Luu{-}Thuy Nguyen. 2019.
\newblock \href {http://arxiv.org/abs/1911.09339} {Emotion recognition for vietnamese social media text}.
\newblock \emph{CoRR}, abs/1911.09339.

\bibitem[{Hochreiter and Schmidhuber(1997)}]{lstm}
Sepp Hochreiter and J\"{u}rgen Schmidhuber. 1997.
\newblock \href {https://doi.org/10.1162/neco.1997.9.8.1735} {Long short-term memory}.
\newblock \emph{Neural Comput.}, 9(8):1735–1780.

\bibitem[{Joulin et~al.(2017)Joulin, Grave, Bojanowski, and Mikolov}]{joulin-etal-2017-bag}
Armand Joulin, Edouard Grave, Piotr Bojanowski, and Tomas Mikolov. 2017.
\newblock \href {https://aclanthology.org/E17-2068} {Bag of tricks for efficient text classification}.
\newblock In \emph{Proceedings of the 15th Conference of the {E}uropean Chapter of the Association for Computational Linguistics: Volume 2, Short Papers}, pages 427--431, Valencia, Spain. Association for Computational Linguistics.

\bibitem[{Kim(2014)}]{kim-2014-convolutional}
Yoon Kim. 2014.
\newblock \href {https://doi.org/10.3115/v1/D14-1181} {Convolutional neural networks for sentence classification}.
\newblock In \emph{Proceedings of the 2014 Conference on Empirical Methods in Natural Language Processing ({EMNLP})}, pages 1746--1751, Doha, Qatar. Association for Computational Linguistics.

\bibitem[{Levenshtein et~al.(1966)}]{levenshtein1966binary}
Vladimir~I Levenshtein et~al. 1966.
\newblock Binary codes capable of correcting deletions, insertions, and reversals.
\newblock In \emph{Soviet physics doklady}, volume~10, pages 707--710. Soviet Union.

\bibitem[{Ljube{\v s}i{\'c} et~al.(2017)Ljube{\v s}i{\'c}, Erjavec, Mili{\v c}evi{\'c}, and Samard{\v z}i{\'c}}]{Croatian-2017}
Nikola Ljube{\v s}i{\'c}, Toma{\v z} Erjavec, Maja Mili{\v c}evi{\'c}, and Tanja Samard{\v z}i{\'c}. 2017.
\newblock \href {http://hdl.handle.net/11356/1170} {Croatian twitter training corpus {ReLDI}-{NormTagNER}-hr 2.0}.
\newblock Slovenian language resource repository {CLARIN}.{SI}.

\bibitem[{Ljube{\v{s}}i{\'c} et~al.(2017)Ljube{\v{s}}i{\'c}, Erjavec, Mili{\v{c}}evi{\'c}, and Samard{\v{z}}i{\'c}}]{Serbian-2017}
Nikola Ljube{\v{s}}i{\'c}, Toma{\v{z}} Erjavec, Maja Mili{\v{c}}evi{\'c}, and Tanja Samard{\v{z}}i{\'c}. 2017.
\newblock Serbian twitter training corpus reldi-normtagner-sr 2.0.

\bibitem[{Luu et~al.(2021)Luu, Nguyen, and Nguyen}]{vihsd}
Son~T. Luu, Kiet~Van Nguyen, and Ngan Luu-Thuy Nguyen. 2021.
\newblock A large-scale dataset for hate speech detection on vietnamese social media texts.
\newblock In \emph{Advances and Trends in Artificial Intelligence. Artificial Intelligence Practices}, pages 415--426, Cham. Springer International Publishing.

\bibitem[{Makhija et~al.(2020)Makhija, Kumar, and Gupta}]{makhija-etal-2020-hinglishnorm}
Piyush Makhija, Ankit Kumar, and Anuj Gupta. 2020.
\newblock \href {https://doi.org/10.18653/v1/2020.coling-industry.13} {hinglish{N}orm - a corpus of {H}indi-{E}nglish code mixed sentences for text normalization}.
\newblock In \emph{Proceedings of the 28th International Conference on Computational Linguistics: Industry Track}, pages 136--145, Online. International Committee on Computational Linguistics.

\bibitem[{Muller et~al.(2019)Muller, Sagot, and Seddah}]{muller-etal-2019-enhancing}
Benjamin Muller, Benoit Sagot, and Djam{\'e} Seddah. 2019.
\newblock \href {https://doi.org/10.18653/v1/D19-5539} {Enhancing {BERT} for lexical normalization}.
\newblock In \emph{Proceedings of the 5th Workshop on Noisy User-generated Text (W-NUT 2019)}, pages 297--306, Hong Kong, China. Association for Computational Linguistics.

\bibitem[{Nguyen and Tuan~Nguyen(2020)}]{nguyen-tuan-nguyen-2020-phobert}
Dat~Quoc Nguyen and Anh Tuan~Nguyen. 2020.
\newblock \href {https://doi.org/10.18653/v1/2020.findings-emnlp.92} {{P}ho{BERT}: Pre-trained language models for {V}ietnamese}.
\newblock In \emph{Findings of the Association for Computational Linguistics: EMNLP 2020}, pages 1037--1042, Online. Association for Computational Linguistics.

\bibitem[{Nguyen et~al.(2021)Nguyen, Rosseel, and Grieve}]{nguyen-etal-2021-learning}
Dong Nguyen, Laura Rosseel, and Jack Grieve. 2021.
\newblock \href {https://doi.org/10.18653/v1/2021.naacl-main.50} {On learning and representing social meaning in {NLP}: a sociolinguistic perspective}.
\newblock In \emph{Proceedings of the 2021 Conference of the North American Chapter of the Association for Computational Linguistics: Human Language Technologies}, pages 603--612, Online. Association for Computational Linguistics.

\bibitem[{Nguyen et~al.(2023)Nguyen, Pham, Le, Luong, Tran, Man, Nguyen, Luu, Nguyen, Bui, Phung, and Nguyen}]{vinai}
Thien~Hai Nguyen, Thinh Pham, Khoi~Minh Le, Manh Luong, Nguyen~Luong Tran, Hieu Man, Dang~Minh Nguyen, Tuan~Anh Luu, Thien~Huu Nguyen, Hung Bui, Dinh Phung, and Dat~Quoc Nguyen. 2023.
\newblock \href {https://doi.org/10.1145/3581754.3584159} {A vietnamese spelling correction system}.
\newblock In \emph{Companion Proceedings of the 28th International Conference on Intelligent User Interfaces}, IUI '23 Companion, page 158–161, New York, NY, USA. Association for Computing Machinery.

\bibitem[{Nguyen et~al.(2015)Nguyen, Nguyen, and Snasel}]{nguyen2015normalization}
Vu~H Nguyen, Hien~T Nguyen, and Vaclav Snasel. 2015.
\newblock Normalization of vietnamese tweets on twitter.
\newblock In \emph{Intelligent Data Analysis and Applications: Proceedings of the Second Euro-China Conference on Intelligent Data Analysis and Applications, ECC 2015}, pages 179--189. Springer.

\bibitem[{Nguyen et~al.(2016)Nguyen, Nguyen, and Snasel}]{nguyen2016text}
Vu~H Nguyen, Hien~T Nguyen, and Vaclav Snasel. 2016.
\newblock Text normalization for named entity recognition in vietnamese tweets.
\newblock \emph{Computational social networks}, 3:1--16.

\bibitem[{Plank et~al.(2020)Plank, Jensen, and van~der Goot}]{plank-etal-2020-dan}
Barbara Plank, Kristian~N{\o}rgaard Jensen, and Rob van~der Goot. 2020.
\newblock \href {https://doi.org/10.18653/v1/2020.coling-main.583} {{D}a{N}+: {D}anish nested named entities and lexical normalization}.
\newblock In \emph{Proceedings of the 28th International Conference on Computational Linguistics}, pages 6649--6662, Barcelona, Spain (Online). International Committee on Computational Linguistics.

\bibitem[{Sennrich et~al.(2016)Sennrich, Haddow, and Birch}]{sennrich-etal-2016-neural}
Rico Sennrich, Barry Haddow, and Alexandra Birch. 2016.
\newblock \href {https://doi.org/10.18653/v1/P16-1162} {Neural machine translation of rare words with subword units}.
\newblock In \emph{Proceedings of the 54th Annual Meeting of the Association for Computational Linguistics (Volume 1: Long Papers)}, pages 1715--1725, Berlin, Germany. Association for Computational Linguistics.

\bibitem[{Sidorenko(2019)}]{sidorenko2019sentiment}
Wladimir Sidorenko. 2019.
\newblock Sentiment analysis of german twitter.
\newblock \emph{arXiv preprint arXiv:1911.13062}.

\bibitem[{Tran et~al.(2022)Tran, Le, and Nguyen}]{bartpho}
Nguyen~Luong Tran, Duong~Minh Le, and Dat~Quoc Nguyen. 2022.
\newblock {BARTpho: Pre-trained Sequence-to-Sequence Models for Vietnamese}.
\newblock In \emph{Proceedings of the 23rd Annual Conference of the International Speech Communication Association}.

\bibitem[{van~der Goot(2019{\natexlab{a}})}]{van-der-goot-2019-monoise}
Rob van~der Goot. 2019{\natexlab{a}}.
\newblock \href {https://doi.org/10.18653/v1/P19-3032} {{M}o{N}oise: A multi-lingual and easy-to-use lexical normalization tool}.
\newblock In \emph{Proceedings of the 57th Annual Meeting of the Association for Computational Linguistics: System Demonstrations}, pages 201--206, Florence, Italy. Association for Computational Linguistics.

\bibitem[{van~der Goot(2019{\natexlab{b}})}]{van2019normalization}
Rob van~der Goot. 2019{\natexlab{b}}.
\newblock \href {https://pure.rug.nl/ws/portalfiles/portal/78256480/Complete_thesis.pdf} {\emph{Normalization and parsing algorithms for uncertain input}}.
\newblock Ph.D. thesis, University of Groningen.

\bibitem[{van~der Goot et~al.(2020)van~der Goot, Ramponi, Caselli, Cafagna, and De~Mattei}]{van-der-goot-etal-2020-norm}
Rob van~der Goot, Alan Ramponi, Tommaso Caselli, Michele Cafagna, and Lorenzo De~Mattei. 2020.
\newblock \href {https://aclanthology.org/2020.lrec-1.769} {Norm it! lexical normalization for {I}talian and its downstream effects for dependency parsing}.
\newblock In \emph{Proceedings of the Twelfth Language Resources and Evaluation Conference}, pages 6272--6278, Marseille, France. European Language Resources Association.

\bibitem[{van~der Goot et~al.(2021)van~der Goot, Ramponi, Zubiaga, Plank, Muller, San Vicente~Roncal, Ljube{\v{s}}i{\'c}, {\c{C}}etino{\u{g}}lu, Mahendra, {\c{C}}olako{\u{g}}lu, Baldwin, Caselli, and Sidorenko}]{van-der-goot-etal-2021-multilexnorm}
Rob van~der Goot, Alan Ramponi, Arkaitz Zubiaga, Barbara Plank, Benjamin Muller, I{\~n}aki San Vicente~Roncal, Nikola Ljube{\v{s}}i{\'c}, {\"O}zlem {\c{C}}etino{\u{g}}lu, Rahmad Mahendra, Talha {\c{C}}olako{\u{g}}lu, Timothy Baldwin, Tommaso Caselli, and Wladimir Sidorenko. 2021.
\newblock \href {https://doi.org/10.18653/v1/2021.wnut-1.55} {{M}ulti{L}ex{N}orm: A shared task on multilingual lexical normalization}.
\newblock In \emph{Proceedings of the Seventh Workshop on Noisy User-generated Text (W-NUT 2021)}, pages 493--509, Online. Association for Computational Linguistics.

\bibitem[{van~der Goot et~al.(2018)van~der Goot, van Noord, and van Noord}]{van-der-goot-etal-2018-taxonomy}
Rob van~der Goot, Rik van Noord, and Gertjan van Noord. 2018.
\newblock \href {https://aclanthology.org/L18-1109} {A taxonomy for in-depth evaluation of normalization for user generated content}.
\newblock In \emph{Proceedings of the Eleventh International Conference on Language Resources and Evaluation ({LREC} 2018)}, Miyazaki, Japan. European Language Resources Association (ELRA).

\bibitem[{Van~Dinh et~al.(2022)Van~Dinh, Luu, and Nguyen}]{vispam}
Co~Van~Dinh, Son~T. Luu, and Anh Gia-Tuan Nguyen. 2022.
\newblock Detecting spam reviews on vietnamese e-commerce websites.
\newblock In \emph{Intelligent Information and Database Systems}, pages 595--607, Cham. Springer International Publishing.

\bibitem[{Vaswani et~al.(2017)Vaswani, Shazeer, Parmar, Uszkoreit, Jones, Gomez, Kaiser, and Polosukhin}]{vaswani2017attention}
Ashish Vaswani, Noam Shazeer, Niki Parmar, Jakob Uszkoreit, Llion Jones, Aidan~N Gomez, {\L}ukasz Kaiser, and Illia Polosukhin. 2017.
\newblock Attention is all you need.
\newblock \emph{Advances in neural information processing systems}, 30.

\bibitem[{Vu et~al.(2018)Vu, Nguyen, Nguyen, Dras, and Johnson}]{vu-etal-2018-vncorenlp}
Thanh Vu, Dat~Quoc Nguyen, Dai~Quoc Nguyen, Mark Dras, and Mark Johnson. 2018.
\newblock \href {https://doi.org/10.18653/v1/N18-5012} {{V}n{C}ore{NLP}: A {V}ietnamese natural language processing toolkit}.
\newblock In \emph{Proceedings of the 2018 Conference of the North {A}merican Chapter of the Association for Computational Linguistics: Demonstrations}, pages 56--60, New Orleans, Louisiana. Association for Computational Linguistics.

\bibitem[{Yang and Eisenstein(2013)}]{yang-eisenstein-2013-log}
Yi~Yang and Jacob Eisenstein. 2013.
\newblock \href {https://aclanthology.org/D13-1007} {A log-linear model for unsupervised text normalization}.
\newblock In \emph{Proceedings of the 2013 Conference on Empirical Methods in Natural Language Processing}, pages 61--72, Seattle, Washington, USA. Association for Computational Linguistics.

\bibitem[{Zupan et~al.(2019)Zupan, Ljubešić, and Erjavec}]{pos}
Katja Zupan, Nikola Ljubešić, and Tomaž Erjavec. 2019.
\newblock \href {https://doi.org/10.1017/S1351324919000366} {How to tag non-standard language: Normalisation versus domain adaptation for slovene historical and user-generated texts}.
\newblock \emph{Natural Language Engineering}, 25:651--674.

\end{thebibliography}

\appendix
\onecolumn

\section{Most Commonly Normalized Words in \textsc{ViLexNorm}} \label{sec:appendixa}

To identify words commonly substituted with their variants, we investigated the ten most frequently occurring 1-syllable and 2-syllable normalized words along with their frequencies and respective variants (refer to Table~\ref{tab:1common-normalized} and \ref{tab:2common-normalized}).

\renewcommand{\arraystretch}{1.4}
\begin{table*}[h]
\centering
\begin{tabular}{p{3.5cm}rp{9.2cm}} 
\hline
{\textbf{1-syllable}} & {\textbf{Distribution}} & \multicolumn{1}{c}{\textbf{Variants}} \\ 
\hline
không (\textit{no}) & 27\% & k, hong, hông, ko, hỏng, kg, hok, hem, khum, hổng, kh, khong, hk, hõng, hog, khom, honggggggg, hogg, khun, hẻm, khưm, k-ko, ứ, khôm, hum, o, hơm, khummm, 0, honk, hỏk, hăm, hongg, kô, khumm, hongggg, hôngggg, hỏg, hôk, ko, hg, khoonng, khôg, khoeng, khok, hôn, khônh, kog, kó, ki, hoq, hônnn, hống \\
tôi (\textit{me}) & 16\% & t, toi, tuôi, toy, toai, tôy, tui, toyy, toii, tôiiiiii, tao, tuiii \\
được (\textit{okay}) & 6\% & đc, dc, dk, đựt, đượt, đk, đx, đươic, đượttt, đươc, ddc, đuọce, đuọc, duoc, đuoc, đực, đfc, dcd, đượtttt, đv, đượk, duocc \\
rồi (\textit{already}) & 4\% & ồi, r, oy, ròi, òii, gòi, roi, ời, gồi, rùi, òi, roy, zòyyyy, ròiii, ùi, roài, rồu, gòy, gùi, rồy, rùiii, gòiii, zòi, roàiii, rồiii, dòii, rầu, roii, goy, rôi, ui, dồi, rui, dồii, gòyy, ròy, roiif, dzồi, rùiiii, dòi, rồiiiii, ròii, ròy, royyy, rùii, rrrrr, gồy, ruid, òy \\
vậy (\textit{so}) & 4\% & dẫy, zậy, v, z, dậy, dị, vậyyy, dzị, vị, d, zị, vay, dzậy, dzi, dạ, dzọ, zay, dãy, zịk, dzayy, dợ, zẫy, zạy, dọ, dì, zz, vạiiii, vầy, zayyy, vạy, vậii, zạyy, vại, dzạy \\
em (\textit{you/he/she}) & 3\% & e, emk, iêm, iêmmmm, iem, ẻm, emm, eim, eng, kem, êm, 3m, êmmmm \\
người (\textit{person}) & 2\% & ng, ngừi, ngta, nguoi, ngừiii, nguời, ny, ngừ, n, ngườii, ngừoi \\
mày (\textit{you}) & 2\% & m, mài, mài, may, màiii, mèy, m \\
với (\textit{with}) & 2\% & vs, zới, dứi, dí, dới, vứi, dzới, zí, vz, zs, vớiiii, vưới, vớii, v, zdí, zúii, w, zứi, voi, va, dứ \\
anh (\textit{you/he}) & 2\% & ank, a, ânh, ah, an, ăng, ann\\
\hline
\end{tabular}
\caption{The most commonly 1-syllable normalized words in \textsc{ViLexNorm} along with their respective distribution percentages and variants.}
\label{tab:1common-normalized}
\end{table*}

\newpage

\begin{table*}[h]
\centering
\begin{tabular}{p{3.5cm}rp{9.2cm}} 
\hline
{\textbf{2-syllable}} & {\textbf{Distribution}} & \multicolumn{1}{c}{\textbf{Variants}} \\ 
\hline
người ta (\textit{people}) & 9\% & ngta, nta, ngt \\
người yêu (\textit{lover}) & 9\% & ny, ngyeu, ngyo, ngiu, ngy, ng iu, ngừi iu, any, eo, ngừi eo, ngyêu, ngêu, nyêu \\
mọi người (\textit{everyone}) & 6\% & mn, mngg, mng, m.ng, mụi ng, m.n, m.n, mậu ngừ, mụi ngừi, mụi ngườiiii \\
nhưng mà (\textit{but}) & 5\% & nhma, nma, nhmà, nhm, nmà \\
anh em (\textit{brothers}) & 3\% & ae, a e \\
bình thường (\textit{normal}) & 2\% & bthf, bthg, bt, bth, binh thuong, bthuong, binh thukng \\
gia đình (\textit{family}) & 2\% & gđ, gd \\
điện thoại (\textit{phone}) & 2\% & dthoai, đth, đt, dt \\
sinh nhật (\textit{birthday}) & 2\% & sn, snhat, xưn nhựt \\
bao giờ (\textit{whenever}) & 2\% & baoh, bg, bh, bjo, bgio \\
\hline
\end{tabular}
\caption{The most commonly 2-syllable normalized words in \textsc{ViLexNorm} along with their respective distribution percentages and variants.}
\label{tab:2common-normalized}
\end{table*}

\newpage

\section{Error Analysis}

To explore the linguistic challenges posed by the lexical normalization task, we examined BARTpho\textsubscript{syllable}'s prediction failures on the development set. Astonishing results were observed, highlighting the model's difficulty in handling the usage of dialects and slang words on social media platforms. This reaffirms the diverse linguistic practices employed by Vietnamese speakers online. Our system also struggled with obfuscated words, a persistent issue in offensive language detection. Furthermore, we encountered instances of word-choice ambiguity. Refer to Table~\ref{err_analysis} for detailed examples and discussion. Importantly, all of these error cases involve intentional spelling variations, thus reinforcing the core objective of our research: to encompass the deliberate linguistic variations prevalent in social media usage.

\renewcommand{\arraystretch}{1.4}

\begin{longtable}{>{\hspace{2pt}}m{0.03\linewidth}>{\hspace{0pt}}m{0.52\linewidth}>{\hspace{0pt}}m{0.38\linewidth}} 
\hline
\multicolumn{1}{>{\centering\hspace{2pt}}m{0.03\linewidth}}{\begin{sideways}\end{sideways}} & \multicolumn{1}{>{\centering\hspace{0pt}}m{0.52\linewidth}}{\textbf{Examples}} & \multicolumn{1}{>{\centering\arraybackslash\hspace{0pt}}m{0.38\linewidth}}{\textbf{Discuss}} \\*
\hline
\multirow{2}{*}[-6ex]{\rotatebox[origin=c]{90}{\textbf{Dialect writing}}} & \textit{Original}:\par{}\colorbox{yellow}{Dìa} Bình định \colorbox{yellow}{mă} \colorbox{yellow}{hống} \colorbox{yellow}{uống} cà phơ là sai lầm nhá\par{}\textit{Ground-truth}:\par{}\colorbox{yellow}{Về} Bình định \colorbox{yellow}{mà} \colorbox{yellow}{không} \colorbox{yellow}{uống} cà phê là sai lầm nhá\par{}\textit{BARTpho\textsubscript{syllable} predicted}:\par{}\colorbox{yellow}{Dìa} Bình định \colorbox{yellow}{mă} \colorbox{yellow}{hống} \colorbox{yellow}{á\'{c}ng} cà phê là sai lầm nhá\par{}(\textbf{English}: \textit{Visiting Binh Dinh without drinking coffee is a mistake}) & The model did not recognize words written in the phonetic accent of Central Vietnamese ("Dìa," "mă," "hống"). Consequently, it retained these words without normalization and incorrectly normalized a canonical word ("uống"). \\* 
\cdashline{2-3}
 & \textit{Original}:\par{}Quả bạn nhiệt tình gửi đ gì cũng cợt \colorbox{yellow}{nhạ}\par{}\textit{Ground-truth}:\par{}Quả bạn nhiệt tình gửi đéo gì cũng cợt \colorbox{yellow}{nhả}\par{}\textit{BARTpho\textsubscript{syllable} predicted}:\par{}Quả bạn nhiệt tình gửi đéo gì cũng cợt \colorbox{yellow}{nhại}\par{}(\textbf{\textbf{English}}:~\textit{An enthusiastic friend, sent anything will joke"}) & A similar mistake was observed with the syllable "nhạ" that~\textit{BARTpho\textsubscript{syllable}} incorrectly chose "nhại" to replace instead of "nhả".\\* 
\hline
\multirow{2}{*}[-8ex]{\rotatebox[origin=c]{90}{\textbf{Slang words}}} & \textit{Original}:\par{}em \colorbox{yellow}{bừn} \colorbox{yellow}{tữn} \colorbox{yellow}{ngke} \colorbox{yellow}{ăng} lóy \colorbox{yellow}{i} \colorbox{yellow}{mò}\par{}\textit{Ground-truth}:\par{}em \colorbox{yellow}{bình} \colorbox{yellow}{tĩnh} \colorbox{yellow}{nghe} \colorbox{yellow}{anh} nói \colorbox{yellow}{đi} \colorbox{yellow}{mà}\par{}\textit{BARTpho\textsubscript{syllable} predicted}:\par{}em \colorbox{yellow}{bừn} \colorbox{yellow}{thiếp} \colorbox{yellow}{người ta} \colorbox{yellow}{ăn} nói \colorbox{yellow}{đi i} \colorbox{yellow}{mò}\par{}(\textbf{\textbf{English}}:~\textit{please stay calm and listen to me}) & In this case, the model struggled with out-of-vocabulary slang words, leading to the selection of incorrect normalized counterparts. \\* 
\cdashline{2-3}
 & \textit{Original}:\par{}mai mốt hong có giành ăn zị nha \colorbox{yellow}{hôn}\par{}\textit{Ground-truth}:\par{}mai mốt không có giành ăn vậy nha \colorbox{yellow}{không}\par{}\textit{BARTpho\textsubscript{syllable} predicted}:\par{}mai mốt không có giành ăn vậy nha \colorbox{yellow}{hôn}\par{}(\textbf{\textbf{English}}:~\textit{don't compete for food like that in the future, okay?}) & Conversely, BARTpho\textsubscript{syllable} failed to identify the slang term "hôn" due to its presence in the formal vocabulary with a different meaning. \\* 
\hline
\multirow{2}{*}[-4ex]{\rotatebox[origin=c]{90}{\textbf{Obfuscated words}}} & \textit{Original}:\par{}Mai lại có gỏi gà dưa hấu , sầu riêng thì \colorbox{yellow}{t.o.i}\par{}\textit{Ground-truth}:\par{}Mai lại có gỏi gà dưa hấu , sầu riêng thì \colorbox{yellow}{toi}\par{}\textit{BARTpho\textsubscript{syllable} predicted}:\par{}Mai lại có gỏi gà dưa hấu, sầu riêng thì \colorbox{yellow}{tôi.o.i}\par{}(\textbf{\textbf{English}}:~\textit{I'm dead with the idea of watermelon-chicken and durian-chicken salad}) & The deliberate separation of characters in the word "toi" (\textit{dead}) using dots caused confusion for the model during normalization. \\* 
\cdashline{2-3}
 & \textit{Original}:\par{}Suy nghĩ của mấy con \colorbox{yellow}{thieunang} khó hiểu lắm\par{}\textit{Ground-truth}:\par{}Suy nghĩ của mấy con \colorbox{yellow}{thiểu năng} khó hiểu lắm\par{}\textit{BARTpho\textsubscript{syllable} predicted}:\par{}Suy nghĩ của mấy con \colorbox{yellow}{thieunang} khó hiểu lắm\par{}(\textbf{\textbf{English}}:~\textit{The thoughts of retarded guys are very hard to get}) & Likewise, intentionally omitting the space between two syllables and diacritics of the word "thiểu năng" (\textit{retarded}) has fooled our system. \\* 
\hline
\multirow{2}{*}[-5ex]{\rotatebox[origin=c]{90}{\textbf{Word ambiguity}}} & \textit{Original}:\par{}Khổ thân mấy con gà,mai~\colorbox{yellow}{m} làm con ăn đã\par{}\textit{Ground-truth}:\par{}Khổ thân mấy con gà,mai \colorbox{yellow}{mình} làm con ăn đã\par{}\textit{BARTpho\textsubscript{syllable} predicted}:\par{}Khổ thân mấy con gà,mai \colorbox{yellow}{mày} làm con ăn đã\par{}(\textbf{\textbf{English}}:~\textit{Poor chickens, tomorrow I will eat one}) & This example highlights the BARTpho\textsubscript{syllable}'s challenge in accurately selecting the appropriate pronoun. In particular, it chose "mày," a second-person pronoun, instead of the correct normalization "mình," which is a first-person pronoun. \\* 
\cdashline{2-3}
 & \textit{Original}:\par{}Chả hiểu sao mình vẫn sống được đến \colorbox{yellow}{bh} nhỉ\par{}\textit{Ground-truth}:\par{}Chả hiểu sao mình vẫn sống được đến \colorbox{yellow}{bây giờ} nhỉ\par{}\textit{BARTpho\textsubscript{syllable} predicted}:\par{}Chả hiểu sao mình vẫn sống được đến \colorbox{yellow}{bao giờ} nhỉ\par{}(\textbf{\textbf{English}}:~\textit{I don't know how I can still be alive until now}) & In another case of ambiguity, the model incorrectly used "bao giờ" (whenever) instead of "bây giờ" (now), illustrating its struggle in distinguishing relative-time words. \\
\hline
\caption{Challenging instances in the Development set from \textsc{ViLexNorm} for the BARTpho\textsubscript{syllable} model.}
\label{err_analysis}
\end{longtable}

\newpage
\section{Extrinsic Experimental Settings} \label{sec:appendixc}

\renewcommand{\arraystretch}{1.4}
\begin{table}[h]
\centering
\begin{tabular}{llr} 
\hline
\multirow{6}{*}{\textbf{TextCNN, BiLSTM, GRU }} & \textit{Training epochs} & 40 \\
 & \textit{Learning rate} & 1e-4 \\
 & \textit{Optimizer} & Adam \\
 & \textit{Loss function} & CrossEntropy \\
 & \textit{Embeddings} & FastText \cite{joulin-etal-2017-bag} \\
 & \textit{Batch size} & 256 \\ 
\hline
\multirow{4}{*}{\textbf{PhoBERT}} & \textit{Version} & base\footnotemark \\
 & \textit{Training epochs} & 2 \\
 & \textit{Learning rate} & 5e-5 \\
 & \textit{Sequence length} & 256 \\
 & \textit{Batch size} & 16 \\
\hline
\end{tabular}
\caption{Training settings for the models in the extrinsic evaluation.}
\label{tab:appena}
\end{table}

\footnotetext{PhoBERT\textsubscript{base} is publicly available on \url{https://huggingface.co/vinai/phobert-base}.}

\end{document}